\def\year{2022}\relax
\documentclass[letterpaper]{article} 
\usepackage{aaai22}  
\usepackage{times}  
\usepackage{helvet}  
\usepackage{courier}  
\usepackage[hyphens]{url}  
\usepackage{graphicx} 
\usepackage{natbib}  
\usepackage{caption} 
\DeclareCaptionStyle{ruled}{labelfont=normalfont,labelsep=colon,strut=off} 
\usepackage{algorithm}
\usepackage{algorithmic}

\usepackage{newfloat}
\usepackage{listings}
\floatstyle{ruled}
\newfloat{listing}{tb}{lst}{}
\floatname{listing}{Listing}

\usepackage{times}
\usepackage{epsfig}
\usepackage{graphicx}
\usepackage{amsmath}
\usepackage{amssymb}
\usepackage{enumitem}
\usepackage{times}
\usepackage{bbm}
\usepackage{mathtools}
\usepackage{longtable}

\usepackage{float}
\usepackage{helvet} 
\usepackage{courier} 
\usepackage{float}

\usepackage{amssymb}
\usepackage{xcolor}
\usepackage{booktabs}
\usepackage{breqn}
\usepackage[pagebackref=true,breaklinks=true,colorlinks,bookmarks=false]{hyperref}
\newcommand{\norm}[1]{\left\lVert#1\right\rVert}
\title{Mixing between the Cross Entropy and the Expectation Loss Terms}

\author {
    Barak Battash,\textsuperscript{\rm 1}
    Tamir Hazan, \textsuperscript{\rm 2}
    Lior Wolf, \textsuperscript{\rm 1}
}
\affiliations {
    \textsuperscript{\rm 1} Tel-Aviv University\\
    \textsuperscript{\rm 2} Technion\\
}
\usepackage{bibentry}

\begin{document}

\maketitle

\begin{abstract}
  The cross entropy loss $-\log p(y | x)$, where $p(y | x)$ is the pseudo  probability of the target label $y$ given the data instance $x$, is widely used due to its effectiveness and solid theoretical grounding. However, as training progresses, the loss tends to focus on hard to classify samples, which may prevent the network from obtaining gains in performance. While most work in the field suggest ways to classify hard negatives, we suggest to strategically leave hard negatives behind, in order to focus on misclassified samples with higher probabilities. We show that adding to the optimization goal the expectation loss $1-p(y | x)$, which is a better approximation of the zero-one loss, helps the network to achieve better accuracy.  We, therefore, propose to shift between the two losses during training, focusing more on the expectation loss gradually during the later stages of training. Our experiments show that the new training protocol improves performance across a diverse set of classification domains, including computer vision, natural language processing, tabular data, and sequences. Our code and scripts are available at supplementary.
\end{abstract}

\section{Introduction}
\begin{figure}
\centering
\includegraphics[width=\linewidth]{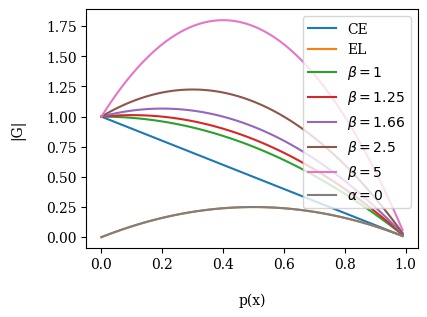}
\vspace{-.3in}
 \caption{The gradients in absolute value for the true label $y$ as a function of $p_y$, for multiple combinations of the CE and EL. When specifying $\beta$, the value of $\alpha$ is 1 and vice versa.}
\label{fig:mixit}
\end{figure}
The Cross Entropy (CE) loss has been dominating the field of training multi-class networks, since the early successes of deep learning a decade ago. It is known to have fast convergence guarantees \cite{hazan2007logarithmic, koren2015fast}, be effective even for 100s of millions of classes~\cite{faces}, and  to be top-k calibrated for any $k\geq 1$~\cite{lapin}. However, CE is a poor approximation of the zero-one loss, which captures the classification accuracy goal, especially for low pseudo-probabilities of the correct class $p(y|x)$ . The CE loss may come in handy at the first stages of training, in which all probabilities are low. However, later on, when most samples are correctly classified, the samples which maintain a low probability may become overly emphasized. This may come at the expense of classifying correctly misclassified samples with higher $p(y|x)$.

Motivated by this concern, we consider an alternative loss, $(1-p(y|x))$. This loss, which we call the expectation loss (EL) is the first order Taylor approximation of the CE. However, for small values of $p(y|x)$, it approximates the zero-one loss much better. Unlike CE, whose gradient is larger for samples with small values of $p(y|x)$, the gradient of EL emphasizes samples that have $p(y|x)=0.5$. 
Furthermore, the EL is the expectation of the zero-one loss, hence, holding much of the zero-one loss properties.

By mixing the two losses, one can emphasize any $p(y|x)$ value between zero and one half in the overall loss term. We propose several recipes to schedule the shifting  of the loss terms. All protocols start with an equal mixture of the two losses. Then, when the CE component has manifested its fast convergence rate, we shift the ratio between the two terms in favor of EL. 

This mixing method is highly effective across multiple benchmarks, including   classification tasks from the UCI tabular datasets~\cite{Bache+Lichman:2013}, CIFAR100
~\cite{Krizhevsky09learningmultiple}, CINIC10~\cite{abs-1810-03505} and ImageNet image recognition problems~\cite{RussakovskyDSKSMHKKBBF14}, five different LSTM configurations on  the Women's Clothing E-Commerce Reviews dataset~\cite{2018arXiv180503687A}, and the application of BERT~\cite{devlin2018BERT} on the GLUE NLP benchmarks~\cite{wang-etal-2018-GLUE}.
Further more, we prove that for mixing the two losses the training procedure holds higher escaping efficiency, thus the training is less likely to stuck in a sharp local minima, and more likely to finish training in a wide better generalizing local minima.
\section{Related work}
Let $x$ denote the input sample and $c$ the number of classes. Let $q(x) \in \mathbb{R}^C$ be the output of the neural network prior to the softmax operator (logits). Let $p(x)=Softmax(q(x))$ be the vector of pseudoprobabilities associated with each of the $C$ labels. Given a label $y=1,2,\dots,C$, The cross entropy loss (CE) can be written as:
\begin{equation}
  L_{ce}(x,y) = -\log p(y | x)\,,
\end{equation}
where the $p(y | x)$ denotes the $y$-th element of the vector $p(x)$, which is the index of the ground truth label. 

A large body of research is devoted to the improvement of CE-based classification in the context of imbalanced data, classification with noisy training labels, and robustness to adversarial samples. However, the literature is much more sparse regarding improving the accuracy obtained with CE.
\newline
\paragraph{\bf Class imbalance} The problem of learning from datasets with an uneven number of training samples is a well-known challenge in machine learning~\cite{Japkowicz2002}, with recent deep learning based approaches~\cite{Huang2016, Dong2018}. Object detection in video and images is known to suffer from this problem due to the high number of possible bounding boxes (background patches) in comparison to the number of positive training samples. This challenge is amplified by the need of object detection methods to achieve a very low false positive rate, while maintaining a high recall.\newline
A modification of the CE loss, called the focal loss~\cite{Tsung-Yi2017} focuses on the set of hard negative samples, by adjusting the CE loss as follows:
\begin{equation}
    L_{focal}(x,y)=-(1-p(y | x))^\gamma \log p(y | x)
\end{equation}
In our experiments, we use $\gamma=2$ and $\alpha=0.25$, which was reported in~\cite{Tsung-Yi2017} to achieve the best performance.
\newline
\newline
\noindent{\bf Classification with noisy labels\quad} The mean absolute error \begin{equation}
L_{mae}= \norm{\vec{y} - p(x)}_1     
\end{equation}
where $\vec{y}$ is the true label as a one hot vector, and, as mentioned above, $p(x)$ is the pseudoprobability vector obtained form the Softmax. This loss has been proposed   as
a noise-robust alternative to the CE loss~\cite{2017arXiv171209482G,janocha2017loss}. However, it showed poor performance for deep neural networks (DNNs)~\cite{2017arXiv171209482G}.  

Veit et al~\cite{Veit} employ a label cleaning network that runs  parallel with the classification network. Two generalized CE losses are suggested: the loss 
$L_q=\frac{1-p(y | x)^q}{q}$
 and its truncated variant.
 
In order to improve robustness to label noise a "tamed" Cross Entropy loss objective (TCE) is formulated~\cite{2018arXiv180507836Z}:
\begin{equation}
    L_{tce}=\frac{1}{1-\alpha} ((1-\log p(y | x))^{1-\alpha}-\frac{1}{1-\alpha})
\end{equation}
This was designed to be identical to CE in the noiseless case, but to exhibit robustness when some of the labels are misleading~\cite{2018arXiv181005075M}.
\newline
\noindent{\bf Robustness to adversarial examples\quad} The vulnerability of DNNs to adversarial attacks first was presented by Goodfellow et al.~\cite{Goodfellow2014}. Subsequent work presented many new attacks~\cite{carlini2017towards,2017arXiv171006081D,Advphysicalworld}, and defences~\cite{KurakinGB16a,2017arXiv170606083M,PapernotM17,PapernotMWJS15}. Recent work \cite{2019arXiv190510626P}, attempted to improve the CE loss by proposing the Max-Mahalanobis center loss
\begin{equation}
    L_{mcc}=\frac{1}{2}\norm{z-\mu^*}_2
\end{equation}
Where $\mu^*$ are the centers of the Max-Mahalanobis distribution and z is the feature vector before the classifier. This work showed a significant improvement in robustness,  even under strong adaptive attack.
\newline
\newline
\noindent{\bf Improving the accuracy of CE\quad} More relevant to our work are contributions which attempt to achieve better accuracy by modifying CE or the training procedure.  The Complement Objective Training (COT) method~\cite{COT2019} employs an objective that leverages information from the complement classes in order to improve both accuracy and robustness to adversarial attacks. The complement objective for one sample is given by:
\begin{equation}
   L_{cot} = -\sum_{j=1, j\not=y}^{N}\left( \frac{p(j | x)}{1-p(y | x)} \right)\log\left( \frac{p(j | x)}{1-p(y | x)}\right)        
\end{equation}

Each training iteration of their method involves two forward-backward passes. The goal of the first pass is to optimize the CE loss. The second's pass goal is to optimize $L_{cot}$. This modified backward pass scheme doubles the number of GFLOPS. In our experiments, we perform a comparison to their method, using the published code.
\newline
 The Maximum Probability based Cross Entropy Loss (MPCE)~\cite{8862886} takes the form:
 \begin{equation}
 L_{mpce} = -(\max_{j} p(j | x) - p(y | x)) \log p(y | x)\,,
 \end{equation}
Unfortunately, we were not able to compare with MPCE, since the authors  did not publish their code, and our attempts to reimplement it resulted in networks that do not converge. We attribute this to the fact that for a large number of classes $c$, for some samples $p(y | x) \approx p(y^* | x) \approx 1/c$, when $y^* = \arg \max_{j } p(\hat j | x)$, which results in $L_{mpce} \approx 0$
\newline
\newline
Modeling SGD as SDE, is a deep-rooted method \cite{walk1992stochastic,kushner2003stochastic} those works derived the first
order stochastic modified equations heuristically. Further, it has been shown by \cite{2015arXiv151106251L} that SGD
can be approximated by an SDE in an  first order weak approximation. 
The first works in the field of studying SGD noise,  approximated SGD by Langevin dynamic with isotropic diffusion coefficient discussed by \cite{sato2014approximation,raginsky2017non,zhang2017hitting}.
\newline
\cite{mandt2017stochastic,2018arXiv180300195Z} made a more accurate SDE formulation, using an anisotropic  noise covariance matrix, our framework includes anisotropic noise, with no need to constraint any property  on the noise covariance matrix.

\section{A Motivating Example}
To study the dynamics of the training process with CE, we trained a  ResNet32~\cite{He2016DeepRL} for 90 epochs on the CIFAR100 dataset~\cite{Krizhevsky09learningmultiple}, which holds 10k test images from 100 classes. On epoch number 70, we began monitoring all the misclassified samples with a softmax probability of less than 0.2 for the ground-truth class. At this stage, out of 10k test images, 6,766 were classified correctly, 611 images were misclassified and had softmax probability of the correct class higher than 0.2, and 2623 were misclassified and had a softmax probability of less than 0.2 .
At the end of the training, we checked how many images were classified correctly. The results shows that 6,482 out of 6,766 ($95.58\%$) images that were correctly classified at epoch 70, were still classified correctly. From the group of misclassified images at epoch 70, which had $p_y>0.2$, 235 ($38.46\%$) images were classified correctly at epoch 90. However, for misclassified images with $p_y<0.2$ at epoch 70, only 88 images out of 2,623 samples ($3.35\%$) were classified correctly at epoch 90.
\newline
Recall that the cross entropy loss (CE) is given by 
$L_{ce}(x,y) = -\log  p(y | x)$, where the $p(y | x)$ denotes the $y$th element of the vector of predicted pseudo-probabilities $p(x)$, given a sample $x$ and a ground truth label $y$. The gradient of CE is given by
\begin{equation}
     \frac{dL_{ce}(x,y)}{dq(j | x)} = \begin{cases}
        p(y | x)-1  & \text{if \(j=y\)} \\
        p(j | x) & \text{if \(j\neq y  \)} \\
\end{cases}
\end{equation}

Therefore, the gradient descent update when training with CE focuses on samples with low $p(y|x)$ and with high $p(j|x)$ for $j\neq y$. However, the results of our motivating experiment indicate that at the end of the training, the samples that are most challenging (low $p(y|x)$) are unlikely to improve.  One may wonder, if a better strategy would be to focus,  instead, on the misclassified samples that are closer to the decision boundary.

\section{Mixing the Expectation Loss with CE}
The EL is given by:
\begin{equation}
  L_{el}(x,y) = 1-p(y | x)\,,
\end{equation}
and its gradient is given by
\begin{equation}
     \frac{dL_{el}(x,y)}{dq(j | x)} = \begin{cases}
        p(y | x)(p(y | x)-1)  & \text{if \(j=y\)} \\
        p(y | x) p(j | x) & \text{if \(j\neq y  \)} \\
\end{cases}
\end{equation}

The maximal value of CE's gradient is one, while EL's is 0.25. Furthermore,   CE's gradient with respect to $q(y | x)$ is maximal at  $p(y | x)=0$, while this gradient  is symmetric around $p(y | x)=0.5$ for EL. The derivative for $q(j | x)$ where $j\neq y$ is maximal for $p(j | x)=1$, in the case of CE, while it is maximal for $p(y | x)=p(j | x)=0.5$ in the EL case. These properties lead to a different training behavior between the two losses, especially for samples for which $p(y | x)$ is close to zero.

Interestingly, the EL is the first order Taylor's approximation of the CE. Using Taylor's remainder theorem for CE, one can verify that in any given interval around $p(y | x)$: 
\begin{equation}
-\log p(y | x) = (1-p(y | x)) + \frac{1}{\xi^2} (1-p(y | x))^2
\end{equation}
for some $\xi$ in an approximated interval. Taylor's remainder theorem emphasizes the disparity in binary classification between three main intervals for the value of $p(y | x)$: (i) the interval $3/4 \le p(y | x) \le 1$ for which the prediction for the label $y$ is correct and the network has high certainty; (ii) the interval $0 \le p(y | x) \le 1/4$ for which the  predicted label is wrong but the network is certain; and (iii) the interval $1/4 \le p(y|x) \le 3/4$ for which the network is uncertain about the training label.
\begin{table}[t]
\centering
\begin{tabular}{@{}l@{~}c@{~}c@{~}c@{}}
\toprule
Model &                        F &CIFAR100 &   CINIC10 \\
\midrule
  ShuffleNetV2 CE&                        - &$ 68.10 \pm 0.08 $& $ 81.35 \pm  0.22$ \\
  ShuffleNetV2 CE+EL &                     0   & $ 68.46 \pm 0.42 $ & $ 81.55\pm 0.16$ \\
  ShuffleNetV2 CE+EL &                  0,0.5 &  $ 68.46 \pm 0.42 $&  $ 81.59\pm0.11$ \\
  ShuffleNetV2 CE+EL &  0-0.5 & $ \textbf{68.51} \pm 0.45 $&  $\mathbf{81.79} \pm0.15 $ \\
  \midrule
      ResNet20 CE&                        - & $ 68.25 \pm 0.53 $ &    $ 83.35\pm 0.10  $\\
   ResNet20 CE+EL &                      0   & $ \textbf{68.58} \pm 0.08 $&  $ 83.53\pm0.14 $\\
   ResNet20 CE+EL &                  0,0.5 &  $ \textbf{68.58} \pm 0.08 $ &   $ 83.53\pm0.14  $\\
   ResNet20 CE+EL &  0-0.5& $ 68.48 \pm 0.14 $ & $\mathbf{83.72} \pm0.06 $\\
   \midrule
      ResNet32 CE&                        -  & $ 69.66 \pm 0.17 $&  $ 84.74 \pm 0.14  $\\
   ResNet32 CE+EL &                        0 &$ \textbf{70.02} \pm 0.21 $&  $84.91 \pm 0.08$ \\
   ResNet32 CE+EL &                  0,0.5 &$ \textbf{70.02} \pm 0.21 $&  $84.96 \pm 0.13 $ \\
   ResNet32 CE+EL &  0-0.5 & $ 69.83 \pm 0.45 $ &  $\mathbf{85.10} \pm 0.10  $\\
   \midrule
   ResNet44 CE&                        - &   $ 70.71 \pm 0.38 $&    $ 84.81 \pm 0.69 $\\
   ResNet44 CE+EL &                        0  & $ \textbf{71.06} \pm 0.45 $&$ 84.99 \pm 0.54 $\\
   ResNet44 CE+EL &                  0,0.5 & $ \textbf{71.06} \pm 0.45 $&$ 85.00 \pm 0.55   $\\
   ResNet44 CE+EL &  0-0.5    & $ 70.93 \pm 0.30 $&  $\mathbf{85.18} \pm 0.50  $ \\   
      \midrule
     ResNet18 CE&                        -&$ 74.54 \pm 0.62 $  &  $ 88.02 \pm 0.24  $ \\
  ResNet18 CE+EL &                        0 &$ 74.58 \pm 0.69 $&  $ 88.22\pm 0.13$ \\
  ResNet18 CE+EL &                  0,0.5 &$ 74.58 \pm 0.69 $&$ \mathbf{88.26}\pm 0.12$ \\
  ResNet18 CE+EL &  0-0.5       & $ \textbf{74.60} \pm 0.57 $ & $88.25 \pm 0.25$ \\
       \midrule
  ResNet34 CE&                        - &$ 74.35 \pm 0.62 $&  $ 88.10 \pm 0.32  $ \\
  ResNet34 CE+EL &                       0 & $\textbf{74.77} \pm 0.72 $ & $ 88.74\pm 0.24$\\
  ResNet34 CE+EL &                  0,0.5  & $\textbf{74.77} \pm 0.72 $&$ 88.77\pm 0.24$\\
  ResNet34 CE+EL &  0-0.5 & $74.69 \pm 1.0$ & $\mathbf{88.85} \pm 0.24$ \\
  \midrule

  PreAct Resnet18 CE&                        -  & $ 74.39 \pm 0.68 $& $ 85.97\pm 1.64 $ \\
  PreAct Resnet18 CE+EL &                        0 &$ \textbf{74.82} \pm 0.36 $&$86.55 \pm 0.54 $ \\
  PreAct Resnet18 CE+EL &                  0,0.5 &$ \textbf{74.82} \pm 0.36 $& $86.60 \pm 0.50$  \\
  PreAct Resnet18 CE+EL &  0-0.5     &$ 74.81 \pm 0.12 $  &  $\mathbf{86.71} \pm0.44 $ \\
     \midrule

  ResNext29 CE&                        - &$ 76.04 \pm 0.64 $&  $ 88.35\pm 0.45 $ \\
  ResNext29 CE+EL &                        0 &$ 76.47 \pm 0.33 $&$88.58 \pm  0.31$\\
  ResNext29 CE+EL &                  0,0.5 &$ 76.47 \pm 0.33 $&  $88.59 \pm 0.31$ \\
  ResNext29 CE+EL &  0-0.5 &$ \textbf{76.52} \pm 0.52 $& $\mathbf{88.72} \pm 0.30$ \\

\bottomrule
\end{tabular}
\caption{Results of eight different models with three and four random seeds on the CINIC10 and CIFAR100 Benchmarks respectively.}
\label{tab:combined}
\end{table}
Taylor's remainder $R_1(p(y | x)) \triangleq \frac{1}{\xi^2} (1-p(y | x))^2$, which accounts for the difference between the CE and EL, varies significantly for the different intervals. For example, when $3/4 \le p(y | x) \le 1$, Taylor's remainder is at most $0.055$, i.e., CE and EL are a good approximation for the interval for which the training prediction for the label $y$ is correct with certainty. On the other hand, when $0 \le p(y | x) \le 1/4$, Taylor's remainder is at least $4.5$, i.e., CE turns out to be a bad approximation for EL. Importantly, in this case, the zero-one loss that measures the true accuracy is significantly more similar to EL than CE, since its distance from EL is $1-p(y | x)$ which is at most $1$, while its distance from CE is at least 3.5. 

Taylor's remainder theorem implies that the expected loss is always upper bounded by the cross entropy loss, namely $1 - p(y | x) \le -\log p(y | x)$. This is a tighter bound than the Pinsker's inequality $(1 - p(y | x))^2 \le - 8 \log p(y | x)$ as it includes second order information.

In a probabilistic point of view, Minimizing $L_{el}$ and minimizing $E[L_{0-1}]$ leads to the same mode estimation, which once again shows the tight connection between the zero-one loss and $L_{el}$.
Although the EL is a better approximation to the zero-one loss, it lacks fast convergence guarantees. In contrast, CE can approach its optimum rapidly, e.g., when considering a single layer network with weight decay. In these cases, stochastic gradient descent methods, such as Adagrad and RMSProp achieve logarithmic regret \cite{mukkamala2017variants}, namely after $T$ optimization steps, the distance between the training loss to the optimal training loss is at most $O(\frac{\log T}{T})$, cf. \cite{shalev2014understanding} Theorem 14.11.   


In our method, we train with a weighted sum of the two losses. Specifically, we employ the loss
\begin{equation}
  L_{\alpha\beta} = \alpha L_{ce} + \beta L_{el}\,,
\end{equation}
where the parameters $\alpha$ and $\beta$ change gradually, such that as training progresses, EL becomes more dominant. 
The gradient of the combined loss is:
\begin{equation}
     \frac{dL_{\alpha\beta}}{dq(x)_j} = \begin{cases}
        \beta p(y | x)^2 +p(y | x)(\alpha-\beta)-\alpha  &\text{if \(j=y\)} \\
         p(j | x)(\beta p(y | x) + \alpha)  &\text{if \(j\neq y \)} \\
\end{cases}
\end{equation}
\begin{table*}[t]
\centering
{%
\begin{tabular}{@{}l@{~}c@{~~}c@{~~}c@{~~}c@{}}
\toprule
            Model &                     CE  & F=0 & F=0,0.5 & F=0-0.5 \\
   \midrule
   ResNet18 &  $69.82 \pm  $ 0.11 &       $ 70.19\pm 0.06 $ &$70.25 \pm  0.08$  &       $ \textbf{70.50} \pm 0.01 $ \\
   ResNet34 &      $ 73.26\pm 0.03 $  &         $73.65 \pm  0.01$  &      $73.91 \pm 0.04  $ & $\textbf{74.07} \pm 0.06$ \\ 
   ResNet50&    $75.74    \pm  0.19$  &            $75.86 \pm 0.01  $  &  $ 76.06 \pm 0.26  $  &        $  \textbf{76.35} \pm 0.02 $ \\
    ResNet101&     $ 77.59 \pm 0.29  $ &  $ 77.90 \pm 0.17 $  &           $78.37 \pm  0.18$  &         $ \textbf{78.51} \pm 0.12$ \\
 SuffleNet &       $ 65.28 \pm 0.07  $  &       $ 65.92 \pm 0.01 $  &         $ 65.51 \pm   0.21  $  &        $ \textbf{65.99} \pm 0.09 $ \\
      ResNeXt50&       $77.30 \pm 0.11 $  &        $ 77.63 \pm 0.03  $  & $ 78.01 \pm 0.02  $ &  $ \textbf{78.25} \pm 0.09$ \\     
   ResNeXt101 &  $ 79.20 \pm 0.00  $  &            $79.45 \pm 0.15 $  &    $ 79.73 \pm 0.00  $ &$  \textbf{79.99} \pm 0.07$ \\   

\bottomrule
\end{tabular}
}
\caption{Top-1 accuracy for seven different ImageNet models, for all four regimes.}
\label{tab:imagenet_short}
\end{table*}
The gradient with respect to $p(y | x)$ has a maximal value at $p(y | x)=0.5(1-\alpha/\beta)$, see Fig.~\ref{fig:mixit}. In order to focus the loss, i.e., obtain maximal response, to a specific probability value $F<0.5$, we, therefore, set
\begin{equation}
\label{eq:beta}
\alpha=1,\quad \beta=(1-2F)^{-1}
\end{equation}
In order to obtain $F=0.5$, we simply set $\alpha=0$ and $\beta=1$.


\subsection{Scheduling}
\label{sec:sched}

Our scheduling method, which determines the weights $\alpha,\beta$, starts in all cases in with equal weights for CE and EL ($F=0$). We examine three different cases:
\begin{enumerate}[leftmargin=0pt,itemindent=1.1em,itemsep=-3pt]
    \item $F=[0]$: In this case, $\alpha=1, \beta=1$ for the entire training and samples with $p(y | x)=0$  will have maximal effect on the gradient.
    \item $F=[0,0.5]$:
    \label{item:startegy2}
    In this
    configuration, we train the model with CE+EL ($\alpha=\beta=1$) for 95\% of the training epochs and in the last 5\% we drop the CE loss and focus on the samples near the decision plane ($\alpha=0, \beta=1$). Hence, in the first part of the training, samples with $p(y | x)=0$  will have maximal effect on the gradient, $p(y | x)=0.5$ on the second part.
    \item $F=[0,0.1,0.2,0.3,0.4,0.5]$:
    \label{item:startegy3}
    where we slowly change our focus, as the training proceeds.  Each intermediate value of $F$ entails a different value of $\beta$ according to Eq.~\ref{eq:beta}, and this value is applied during one phase of training. The phases are of equal length and each constitutes one-sixth of the training epochs. For brevity, we also use the shorthand notation for this configuration and write F=0-0.5. 
\end{enumerate}

\subsection{Escaping Efficiency}
Let us first formulate the stochastic gradient descent update rule as a stochastic differential equation(SDE), the equivalence is widely discussed \cite{}, the SDE is as follows:
\begin{equation}
    dW_t = -\nabla_{W_t}L(W_t)dt + \sqrt{\eta \Sigma_t}dB_t
\end{equation}
Where $\Sigma_t$ is the noise covariance matrix, $\eta$ is the learning rate and $B_t$ is the Brownian motion.The noise covariance matrix is formulated as follows:
\begin{equation}
    \Sigma_t = \frac{1}{m} \left[\frac{1}{N}\sum_{i=1}^{N} \nabla \ell(x_i,W_t) \nabla \ell(x_i,W_t)^T -\nabla L(W_t) \nabla L(W_t)^T \right]
\end{equation}
Where N is the number of samples, and m is the batch size.
The escaping efficiency characterizes the ability to
escape from a local minimum  $W^*$, and is formulated as:
\begin{equation}
    EE \triangleq \mathop{\mathbb{E}_x}[L(W_t) - L(W^*)] 
\end{equation}
An approximation to this measure was derived by \cite{2018arXiv180300195Z}  
  $  EE \approx \frac{t}{2}Tr(H\Sigma)$. 
Our derivation shows that one of the reasons training with $L_{\alpha\beta}$ as an objective, improves the Top1 accuracy is the ability of $L_{\alpha\beta}$ to increase the escaping efficiency, as can be seen:
\begin{align}
  \MoveEqLeft[3] EE(L_{\alpha\beta}) - EE(L_{ce}) > \\{}&
    \left(\beta^3 +\frac{3\beta^2}{M}+\beta\left(\frac{2}{M^2}+\frac{1}{M} \right)\right)Tr\left(F_pF_p^T+2H_pF_p \right)
\end{align}
Where:$F_p \triangleq \mathop{\mathbb{E}_x}[\nabla p(y|x,W_t)\nabla p(y|x,W_t)^T]$,\newline $H_p\triangleq -\mathop{\mathbb{E}_x}[\nabla^2p(y|x,W_t)]$,$M=max_{x\in\mathop{\mathbb{B}}}p(y|x,W_t)$. We refer the reader to the supplementary for the full proof, derivation for the case $\alpha\neq 1$ where it shows that $EE(L_{ce})> EE(L_{el})$, and more elaborations.
It is desirable that the escaping efficiency will be  large such that the training process will be able to rapidly escape from local minima, and globally search for a better basin.
\begin{figure}[t]
    \includegraphics[width=\linewidth]{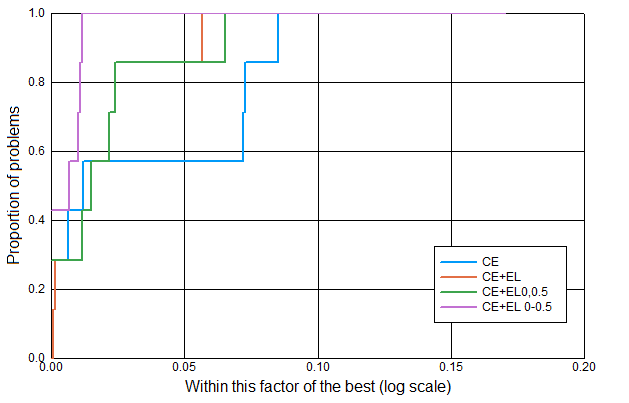}
     \caption{The Dolan-Mor\'{e} profile for the various configurations obtained on the GLUE benchmark. The x axis is a threshold $\tau$ and the y axis is the ratio of experiments for which each of the four mixing configurations obtains a performance that is at least $\tau$ times the maximal accuracy.}
     \label{fig:BERT_dm}
\end{figure}
\section{Experiments}
We have conducted multiple experiments comparing the performance of CE to EL and to a mixture of both. 
We also compared to the focal-loss method~\cite{Tsung-Yi2017}, in the UCI experiments, since UCI consists a large portion of unbalanced datasets, and this is the focal-loss main goal. Lastly, we also compare our method to the COT method~\cite{COT2019}  on three different datasets.

When comparing different methods, we made sure to use the exact same training configuration and hyperparameters. The hyperparameters were chosen as the ones prescribed in various code repositories for the CE loss. 

We employ four types of benchmarks : (i) computer vision datasets, namely, CIFAR100~\cite{Krizhevsky09learningmultiple}, CINIC10~\cite{abs-1810-03505}, and ImageNet~\cite{RussakovskyDSKSMHKKBBF14}, testing various network architectures for each, (ii) NLP problems from the GLUE benchmark~\cite{wang-etal-2018-GLUE}, with the BERT architecture~\cite{devlin2018BERT},  (iii) multi class classification on 
the Womens Clothing E-Commerce Reviews dataset \cite{2018arXiv180503687A} with  two different LSTM-based architectures and (iv) 121 classification problems from the UCI machine learning repository~\cite{Bache+Lichman:2013} using fully connected networks.
\begin{table*}[t]
\centering
\begin{tabular}{lcccccc}
\toprule
        Model &                        F & Arch 1 & Arch 2& Arch 3& Arch 4   & Arch 5 \\
\midrule
     CE &                        - &     $61.08 \pm 0.83$&$60.46 \pm 1.50$&$60.07\pm1.34$&$63.26\pm0.54$ &$63.14\pm0.81$ \\
    CE+EL &                       0 &     $60.81\pm1.30$&$\textbf{61.52} \pm 1.19$&$61.12\pm1.18$&$63.20\pm0.68$ &$\textbf{63.79}\pm1.20$\\
    CE+EL &                  0,0.5 & $60.95\pm1.32$&$\textbf{61.52} \pm 1.19$&$61.12\pm1.18$&$63.20\pm0.68$&$\textbf{63.79}\pm1.20$ \\
    CE+EL &  0-0.5 &     $\textbf{61.57}\pm1.20$&$61.36\pm 1.30$&$\textbf{61.27}\pm1.15$&$\textbf{63.46}\pm0.45$&$63.77\pm0
    71$ \\
\bottomrule
\end{tabular}
\caption{Classification accuracy on the Women's Clothing E-Commerence Reviews dataset for the five LSTM configurations. Architectures 1-3 employ one LSTM layer with an increasing capacity. Architectures 4-5 employ two layers.}
\label{all5archs}
\end{table*}
\newline
In each experiment, except the UCI experiment (where we examine six models), we examine four different loss configurations:  CE loss only, and the three mixtures of EL and CE described above (F=0, F=0,0.5, and F=0-0.5).
The exact hardware and software environment for each experiment will be released together with the pretrained weights. ImageNet experiments were trained on four Nvidia's GPUs, UCI experiments were trained on Intel's CPU, and the rest of the experiments were conducted on a single Nvidia GPU.

The number of repeats per experiment was affected by the required computational resources. The same random seeds were used for all the experiments on each benchmark.  The seeds were chosen arbitrarily. No other seeds were attempted and no cherry picking took place.


\begin{figure*}[t]
 
 \begin{tabular}{cc}
 
 \includegraphics[width=.4975\linewidth]{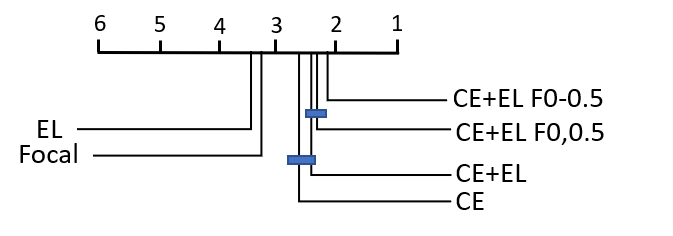}
 
 \includegraphics[width=.4975\linewidth]{Plots/1fc_seed60.png}\\
 \hspace{-3.6in}(a)&\hspace{-1.96in}(b)\\
\includegraphics[width=.4975\linewidth]{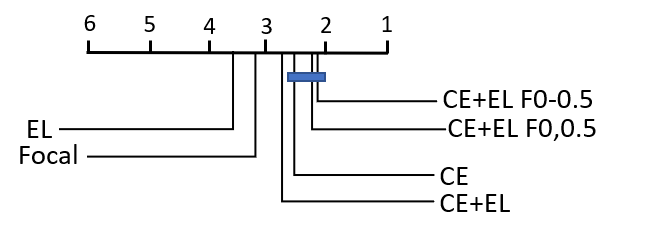}

 \includegraphics[width=.4975\linewidth]{Plots/2fc_seed1234.png}\\

\hspace{-3.6in}(c)&\hspace{-1.96in}(d)\\
\end{tabular}
\caption{ Visualization of post-hoc tests following~\cite{demvsar2006statistical} based on the Conover p-values, further adjusted by the Benjaminyi-Hochberg FDR method. Groups of configurations that are not
significantly different (at p = 0.5) are connected.
 (a) A linear classifier with a random seed, (b) A non-linear classifier with  the same random seed .(c,d) same for a different  a random seed.}
\label{fig:uci_1d}
 \end{figure*}

\subsection{Recognition in CIFAR100 and CINIC10}
\label{sec:cinic10}
We showcase our method on two low-resolution image recognition benchmarks: CIFAR100 \cite{Krizhevsky09learningmultiple}  and CINIC10 \cite{abs-1810-03505}. We trained eight different architectures: ShuffleNetV2~\cite{shufflenet}, ResNet20, ResNet32, ResNet44 \cite{He2016DeepRL}, ResNet18, ResNet34 \cite{He2016DeepRL}, ResNext29 \cite{xie2016aggregated} and PreAct ResNet18 \cite{he2016identity}.

The CIFAR100 dataset holds 50k training images and 10k test images. Each image is of a  32x32 pixel resolution. We trained the network using SGD with a mometum of 0.9 and  a weight decay of 1e-4, for 240 epochs, using an initial learning rate of 0.1, which was decreased at epochs 120, 160 and 200 by a factor of 10. 
We follow the standard data-augmentation techniques, including random cropping, random horizontal flip and normalization. Our reported results are the average of four different seeds, the results for each seed separately are provided in the supplementary.

CINIC-10 includes 180k training-  and 90k test-images with the same spatial resolution as CIFAR100. Training was done using SGD with a mometum of 0.9 and a weight decay of 1e-4 for 90 epochs. The initial learning rate was 0.1, which decreased at epochs 30 and 60 by a factor of 10. The pre-processing is the same as for CIFAR100, with the addition of random rotation for DNNs with more than one million parameters. The reported results are the average of 3 different seeds. See supplementary for the full results.

The results for CIFAR100 and CINIC10 are shown in Tab.~\ref{tab:combined}. In all experiments, just using CE is suboptimal to all other three loss-mixing configurations. In CIFAR100, either the gradual combination ($F=0-0.5$) and the other two types take the top ranking and in CINIC10, in all cases, except one, the gradual combination is preferable.  

Observing Fig.~\ref{fig:mixit}, one could wonder if the gain from our method is the result of increasing of the gradient norm and not the result of shifting the gradient focus.
Hence, we did several experiments, where a usual training session using $L_{ce}$ were used , however we factored  the gradient to have the same gradient norm as training using $L_{\alpha\beta}$ strategy \ref{item:startegy3}. The best result was given by using the following method:
We desire to compare the area under the gradient curve of both of the gradients.
For $L_{ce}$ the volume is simply:
\begin{equation}
V_{ce} = \frac{1}{2}
\end{equation}
For $L_{\alpha\beta}$ we consider two cases: when $i=j$
\begin{equation}
    V_{ce+el} = \int_0^1 \frac{dL_{\alpha\beta}}{dq(x)_j} dp(y|x)
\end{equation}
and when $i \neq j$:
\begin{equation}
    V_{ce+el} = \int_0^1\int_0^1 \frac{dL_{\alpha\beta}}{dq(x)_j} dp(y|x)dp(j|x)
\end{equation}
We then create a new learning loss by multiplying the learning rate for the cross entropy loss by the factor $M = \frac{V_{\alpha+\beta}}{V_{ce}}$.
 The results in the supplementary show that the modified cross entropy method is not competitive.

\subsection{ImageNet}

The experiments were repeated for the ImageNet-2012 classification dataset. The ImageNet dataset contains 1.3 million training images and 50,000 validation images, from a total of 1,000 classes. This dataset is considerably larger than both CIFAR100 and CINIC10 and the images are of much larger resolution. We, therefore, employ deeper networks. We followed the official PyTorch \cite{paszke2017automatic} training script., i.e., we trained our models for 90 epochs, where the learning rate was initially set to 0.1 and is decayed by
a factor of 0.1 at 30 and 60 epoch. The augmentation procedure consists of random resizing and cropping in order to achieve 224 $\times$ 224 resolution and a horizontal flip with a probability of one half. At inference time, only the 224 $\times$ 224 central crop of each image is tested.

We examined the four configurations (as in previous datasets) on seven different models, each with two different seeds. The results are depicted In Tab.~\ref{tab:imagenet_short} (the results per seed are in the supplementary). As can be seen, in all architectures, using the CE loss is the lowest ranking method, and the gradual mixing configuration is ranked first.

\subsection{BERT on the GLUE Benchmark}
We examined our method on seven classification tasks from the GLUE benchmark. 
The models for each task were a finetuned BERT model \cite{devlin2018BERT}\footnote{https://github.com/huggingface/transformers}. 
The GLUE Benchmark differs from the other experiments by having, for example, a low number of classes. The GLUE MNLI task has three classes and the rest of the tasks have two. This may affect the per-class probabilities and, therefore, affect our results.  Furthermore, the focus in GLUE is on finetuning the self-supervised BERT model and not on training from scratch. 
The fine-tuning procedure employed a maximum sequence length of 64, a learning rate of $2e-5$, and trained for six epochs. The evaluation is on the dev set of the benchmark. 
\begin{table*}[t]
\centering
\begin{tabular}{llrrrrrr}
\toprule
&&\multicolumn{3}{c}{Linear classifier} & \multicolumn{3}{c}{Classifier with one hidden layer}\\
\cmidrule(lr){3-5}
\cmidrule(lr){6-8}
            Model &   F & \#Wins  &  $\Delta$Acc & Mean rank& \#Wins  &  $\Delta$Acc & Mean rank\\
\midrule
     CE     &-      &33&0&2.59&26&0&2.70\\
     CE+EL&0        &32&0.09&2.39&30&0.12&2.62\\
     CE+EL&0,0.5&31&0.20&2.30&37&0.27&2.40\\
     CE+EL&0-0.5&\textbf{48}&\textbf{0.30}&\textbf{2.12}&\textbf{42}&\textbf{0.30}&\textbf{2.37}\\
     EL &-          &33&-2.71&3.25&33&-4.73&3.52\\
     Focal &-       &26&-0.56&3.13&25&-0.89&3.03\\
\bottomrule
\end{tabular}
\caption{Summary results on the UCI dataset. The number of wins is the number of datasets for which every method obtained the best performance, out of 121 datasets. $\Delta$Acc stands for the mean improvement in top-1 accuracy over the CE baselines. Mean rank is the average ranking of each method among the six methods. The reported statistics are an average of two runs with two different seeds that are fixed between all methods.}
\label{tab:uci_sum}
\end{table*}
\newline
The reported results are an average of four different seeds (the full results are in the supplementary). A Dolan-Mor\'{e} plot summarizing the results n shown in Fig.~\ref{fig:BERT_dm}. These plots show the ratio of experiments for which each method presents results that are as high as the best method (for this experiment) times a threshold.
\newline
The dominant methods present plots that are above the other methods, since they obtain, for each threshold, a higher ratio of experiments that are within a margin of the best results. As can be seen, the gradual mixing method outperforms the other settings. CE, by itself, has a performance profile that is mostly lower than the other settings.

\subsection {LSTM multi-class classification}

We  examined five different RNN architectures on the Women's Clothing E-Commerce Reviews dataset \cite{2018arXiv180503687A}. This dataset consists of reviews written by
real customers, given an item’s review comment. The goal of the classification is to predict the rating 1-5, where 5 is the best. 
The five architectures all have an embedding layer which maps the input tokens to vectors in $\mathbb{R}^{E_d}$, one or two LSTM layers, each with $H_d$ LSTM cells. Then a linear projection from the output of the LSTM at the end of the sequence is used for classification. The five configurations are:\newline (1) one LSTM layer, $H_d=50,E_d=50$, (2) one LSTM layer, $H_d=75,E_d=75$, (3) one LSTM layer, $H_d=100,E_d=100$, (4) two LSTM layers, $H_d=50,E_d=50$, (5) two LSTM layers, $H_d=100,E_d=100$. %
\newline

Each configuration was examined on eight different seeds and the mean performance as well as the standard deviation is shown in Tab.~\ref{all5archs}. In all cases, except for the forth architecture CE is outperformed by all three mixing protocols. For this architecture, it is outperformed by F=0-0.5.
\newline
\subsection{UCI Datasets}

We examine the performance of shallow neural networks on the 121 classification datasets of the UCI Machine Learning repository, to which the train/test splits were provided by the authors of~\cite{selu}. The size of the datasets ranges between 10 and 130,000 data points and the number of features from 4 to 250. We use two architectures, the first being a linear classifier, which consists of one fully connected layer of size $I \times C$, where $I$ is the input dimension and $C$ is the number of classes. The second architecture examined consists of two fully connected layers with a Relu activation after the first. The first projection matrix has a size of $I \times I$ and the second $I \times C$.
\newline
The models were trained for 50 epoch with a batch of 8, with a sweep on the learning rate (0,01,0.005,0.001), the best for each method was taken based on the validation set. The results of the experiments, at the best epoch which determined by the validation set, as measured on the test set, are given, in full, in the supplementary.
\newline

Tab.~\ref{tab:uci_sum} lists the summary statistics for these experiments, for the two architectures. The reported statistics are: (1) Count of top rank, i.e.,  the number of times each method outperformed or matched all other methods. This does not sum to 121, since multiple models were able to achieve the best accuracy.
(2) Average improvement in top-1 accuracy over the cross entropy model. (3) Mean rank, i.e., the average position in the sorted list of obtained accuracies on all datasets. In this measurement, lower is better.
 \newline

As can be seen, the gradual mixing configuration outperforms the other methods for both architectures. Both other mixing configurations ($F=0$ and $F=0,0.5$) also outperform CE. EL by itself is not competitive, and the focal loss is also outperformed by CE. As expected, the Friedman rank sum test for multiple correlated samples in a two-way balanced complete block design reveals that the methods in each experiment differ at a very low p-value. Fig.~\ref{fig:uci_1d} presents the results of a post-hoc p-value analysis of all possible pairs using the Conover p-value corrected for multiple hypothesis by the Benjaminyi-Hochberg FDR method, following the recommendations of~\cite{demvsar2006statistical}.

 \begin{table}[t]
\centering
\begin{tabular}{@{}llccc@{}}
\toprule
Dataset& Model  &CE &Ours & COT  \\
\midrule
CIFAR100&ResNet110   &70.16&70.24&{\bf 71.36} \\
CIFAR100&PreAct ResNet18   &74.18&74.57&{\bf 75.43} \\
CIFAR100&ResNeXt29&75.88&76.59&{\bf 77.60}\\
\midrule
Tiny INet&ResNet50   &60.61&{\bf 61.67}&60.80 \\
Tiny INet&ResNet101   &61.77&{\bf 63.12}&62.65 \\
\midrule
ImageNet&ResNet50   &75.74*&{\bf 76.35} &75.60 \\
\bottomrule
\end{tabular}
\caption{Comparison with COT \cite{COT2019}  on multiple datasets. Reported is the Top-1 accuracy in percent. For our method, we only tested the F=0-0.5 mixing protocol. $^*$On ImageNet, we were able to obtain a higher baseline value than the 75.30 reported by the previous work~\cite{COT2019}.}
\label{tab:comp_cot}
\end{table}

\subsection {Comparison with COT}
Finally, we compare our results with COT \cite{COT2019} on three different datasets, using the architectures which were used in their paper. The first is dataset is the CIFAR100 dataset, where three models were used: ResNet110, Pre-act-ResNet and ResNext29. The second dataset is Tiny-ImageNet~\cite{wu2017tiny}, in which ResNet50 and ResNet101 were used. Lastly, on ImageNet, we compare using the only model they trained on, which is ResNet50.  

For a fair comparison, we followed the training protocol given in~\cite{COT2019}. Training took place for 200 epochs and the learning rate decreased by a factor of 10 at epochs 100 and 150. The statistics for COT in ImageNet and Tiny-ImageNet are obtained from their publication, while their results on CIFAR100 are re-implemented to verify. This is due to the long training of the first two.
\newline

The results are shown in Tab.~\ref{tab:comp_cot}. Our accuracy values are not identical to those in the previous tables (where there is overlap), due to a change of the training protocol to fit COT's training regime. Evidently, there is a clear advantage for COT on CIFAR100, but not on the larger datasets. It seems that COT acts as an effective regularizer when training large models on a small dataset but does not present a significant advantage on other datasets. We note that while our method does not add GFLOPS to the baseline method, COT increase the GFLOPS by a factor of 2, and as claimed in their paper, the runtime increases by a factor of 1.6.
\newline

\subsection {Mixing EL and CE with extra regularization}

In each of the experiments above weight decay and data augmentations were used, according to the best practices (designed for CR) per benchmark. To evaluate whether the gap in performance observed between the mixed loss and CE is maintained when applying addition regularization, we employ ResNet34 on the CINIC10 dataset, using the same regime mentioned in Sec.~\ref{sec:cinic10}. The list of regularizations applied includes augmentation techniques such as Random Erasing~\cite{zhong1708random} and color jitter, and dropout with probability of $p=0.2$ (higher values of $p$ did not lead to improvement for CE).  The results are presented in Tab.~\ref{tab:more_reg}. As can be seen, there is a gap in performance between the mixed losss and CE across all regularization options.
\begin{table}[t]
\centering
\begin{tabular}{@{}lcc@{}}
\toprule
Regularization & CE & CE+EL (F=0-0.5)\\
\midrule
{Baseline} regularization & 88.10 & {\bf 88.85}\\
Random Erasing & 89.11 & {\bf 89.66}\\
Random Erasing w/o rotation & 89.34 & {\bf 89.82} \\
Color jitter & 87.68 & {\bf 88.81} \\
Dropout & 88.35 & {\bf 88.61}\\
\bottomrule
\end{tabular}
\caption{The effect of regularization on the Top-1 accuracy (\%) as evaluated on the CINIC10 dataset when running a ResNet34. The baseline regularization is the regularization described in Sec.~\ref{sec:cinic10}.}
\label{tab:more_reg}
\end{table}
\newline
\section{Conclusions}

The cross-entropy loss and the expectation loss are similar in  many ways. As shown above, the expectation loss is the first order approximation of the cross entropy loss. Additionally both compare the one-hot distribution and the obtained probabilities. Let $v$ be a zero-one distribution for the true label, i.e., for a training sample $(x,y)$ the zero-one probability is $v_y = 1$ and $v_{\hat y} =0$ for any $\hat y \ne y$. In this case, the cross-entropy loss is given by $KL(v || p)$. The the expectation loss is directly linked to the sum of absolute differences of $p$ and $v$ since $\|v - p \|_1 = 1-p_y + \sum_{\hat y\neq y}p_{\hat y} =  2(1-p_y)$.
Despite these connections, cross-entropy is more effective than the expectation loss and the former is ubiquitous,  while the latter is rarely used. However, we provide motivation, a theoretical support, and an extensive experimental evidence in favor of  mixing of the two losses, at least in the last stages of the training process, when cross-entropy loss leads to a strong focus on the hard negative samples.\newline The gain in performance is consistent across multiple domains despite avoiding a hyperparameter search for the modified loss. We hypothesis that the gain in performance given by mixing the loss terms, is thanks to the higher escaping efficiency, which  allows us to explore a larger parts of
the parameter space without becoming trapped in sharp local minima.  
\newline
\newline
\newline
\newline
{\small
\bibliography{aaai22}
}

\end{document}